\documentclass[10pt,twocolumn,letterpaper]{article}

\usepackage[pagenumbers]{iccv} 

\usepackage{graphicx}
\usepackage{amsmath}
\usepackage{amssymb}
\usepackage{booktabs}

\usepackage{amsmath,amssymb} 
\usepackage{color}
\usepackage{varwidth}
\usepackage{xcolor}
\usepackage{soul}

\usepackage{verbatim}
\usepackage{multirow}
\usepackage{adjustbox}
\usepackage{array}
\usepackage{bbm}
\usepackage{mathtools}
\usepackage{pifont}

\usepackage{scalerel}
\usepackage{listings}
\newcolumntype{x}[1]{>{\centering\let\newline\\\arraybackslash\hspace{0pt}}p{#1}}

\newcommand{\RedX}{\textcolor[HTML]{db3b21}{\ding{55}}}
\newcommand{\GreenCheck}{\textcolor[HTML]{1aaa55}{\ding{51}}}

\usepackage[symbol]{footmisc}

\usepackage{dirtytalk}

\definecolor{iccvblue}{rgb}{0.21,0.49,0.74}
\usepackage[pagebackref,breaklinks,colorlinks,allcolors=iccvblue]{hyperref}

\def\paperID{3940} 
\def\confName{ICCV}
\def\confYear{2025}

\title{DesignLab: Designing Slides Through Iterative Detection and Correction}

\author{
Jooyeol Yun$^{1,2}$, \, Heng Wang$^1$, \, Yotaro Shimose$^1$, \, Jaegul Choo$^2$, \, Shingo Takamatsu$^1$\\
$^1$Sony Group Corporation, \, $^2$Korea Advanced Institute of Science and Technology (KAIST)\\
}
\begin{document}
\twocolumn[{
\renewcommand\twocolumn[1][]{#1}
\maketitle
\begin{center}
    \centering
    \captionsetup{type=figure}
    \includegraphics[width=\textwidth]{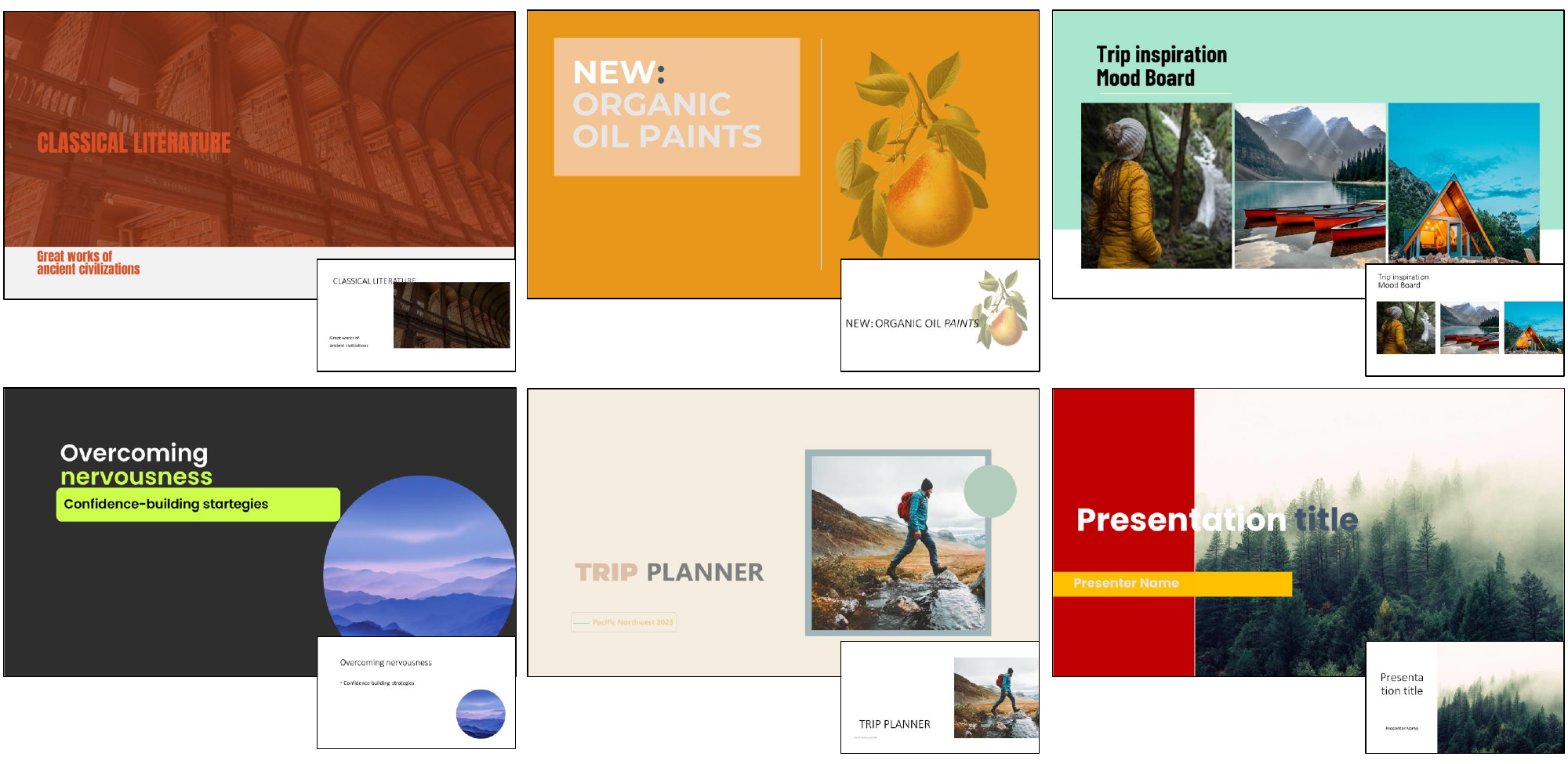}
    \vspace{-0.5cm}
    \captionof{figure}{DesignLab progressively improves initial presentation designs (bottom right) by adding shapes, colors, and text attributes. }
\end{center}
}]

\begin{abstract}
Designing high-quality presentation slides can be challenging for non-experts due to the complexity involved in navigating various design choices. 
Numerous automated tools can suggest layouts and color schemes, yet often lack the ability to refine their own output, which is a key aspect in real-world workflows. 
We propose DesignLab, which separates the design process into two roles, the design reviewer, who identifies design-related issues, and the design contributor who corrects them. 
This decomposition enables an iterative loop where the reviewer continuously detects issues and the contributor corrects them, allowing a draft to be further polished with each iteration, reaching qualities that were unattainable. 
We fine-tune large language models for these roles and simulate intermediate drafts by introducing controlled perturbations, enabling the design reviewer learn design errors and the contributor learn how to fix them.
Our experiments show that DesignLab\footnote[2]{\vspace{-0.8cm}\href{https://yeolj00.github.io/personal-projects/designlab}{https://yeolj00.github.io/personal-projects/designlab}} outperforms existing design-generation methods, including a commercial tool, by embracing the iterative nature of designing which can result in polished, professional slides. 
\end{abstract}


\section{Introduction}
Presentation slides do more than just capture one's attention. 
They become powerful visual tools that anchor a memorable message. 
Yet, for many, the process of creating high-quality slides remains an overwhelming challenge, as it demands a series of nuanced decisions, ranging from content placement to color schemes, typography, and the seamless integration of multimedia elements.
With the sheer volume of design options available, achieving a polished, professional outcome remains daunting for non-experts. 
This presents a fundamental problem that is not simply a matter of investing time, but of navigating the inherent complexity of design itself. 

Automated design tools have attempted to address this challenge by suggesting layouts~\cite{layoutdm, cole, ralf, webrpg} or generating decorative elements~\cite{flexdm, markupdm}. 
While these tools can offer reasonable starting points for creating slides, the result often requires additional editing before they are ready for final use.
More importantly, these approaches fall short in supporting users to iteratively refine the initial output, which is often the most crucial part of the designing process. 
This highlights the need for solutions that foster continuous improvements, rather than simply offering static suggestions. 

In this paper, we tackle a practical scenario where a user starts from an initial rough draft and seeks to refine it into a final design. 
Real-world design processes often revolve around iterative cycles of suggesting, accepting, and rejecting changes, yet prior approaches have largely overlooked this iterative nature of detecting issues and implementing corrections. 
To capture this aspect, we introduce two specialized roles: the design contributor, which modifies specific elements based on requests, and the design reviewer, which identifies elements requiring improvement. 
Our objective is to progressively refine rough drafts over multiple revisions, with each round of feedback addressing remaining issues and driving further enhancement.

Building on the concepts of the design contributor and reviewer, we fine-tune large language models (LLMs) to fill these roles. 
To make presentation slides suitable inputs for LLMs, we convert them into a structured JSON format, capturing elements such as text boxes, images, and layouts. 
Since only complete designs are typically available, we simulate rough drafts to train on pairs of rough designs and their polished version. 
Specifically, we introduce perturbations to the slides, such as altering fonts, shifting alignments, and adjusting colors, so that the perturbed slides resemble imperfect drafts. 
The design reviewer is trained to \emph{detect} these perturbations, learning to identify what is wrong, while the design contributor is trained to \emph{correct} them, understanding how to improve the design. 
By explicitly separating detection (discriminative) from correction (generative), our approach effectively decomposes the distinct cognitive processes required for designing, enabling each role to benefit from specialized training objectives. 
Importantly, this iterative refinement process allows our system to tackle complex design challenges by repeatedly isolating and correcting individual design flaws, rather than merely generating statistical averages of the training data.

We evaluate the effectiveness of our approach using real-world examples of initial presentation drafts in need of improvement.
Our experiments, which include a user study, demonstrate that the decomposed roles of the design contributor and reviewer facilitate progressive improvements, outperforming existing methods that lack support for an iterative process.
Additionally, we showcase an interactive use case of our approach, where users can manually select unsatisfactory elements of the design for enhancement or choose from multiple design candidates to refine the presentation according to their preferences.

Our contributions are threefold:
\begin{itemize}
    \item We introduce DesignLab, the first framework to separate error detection from correction, reflecting real-world design processes where continuous feedback encourages high-quality results.
    \vspace{2pt}
    \item We also showcase an interactive interface, enabled by the decomposed design framework, assisting users to identify and refine unsatisfactory elements. 
    \vspace{2pt}
    \item Our experiments demonstrate that our iterative approach outperforms existing design generation methods, including a commercial tool. 
\end{itemize}


\begin{figure}[t]
    \centering
    \includegraphics[width=0.75\linewidth]{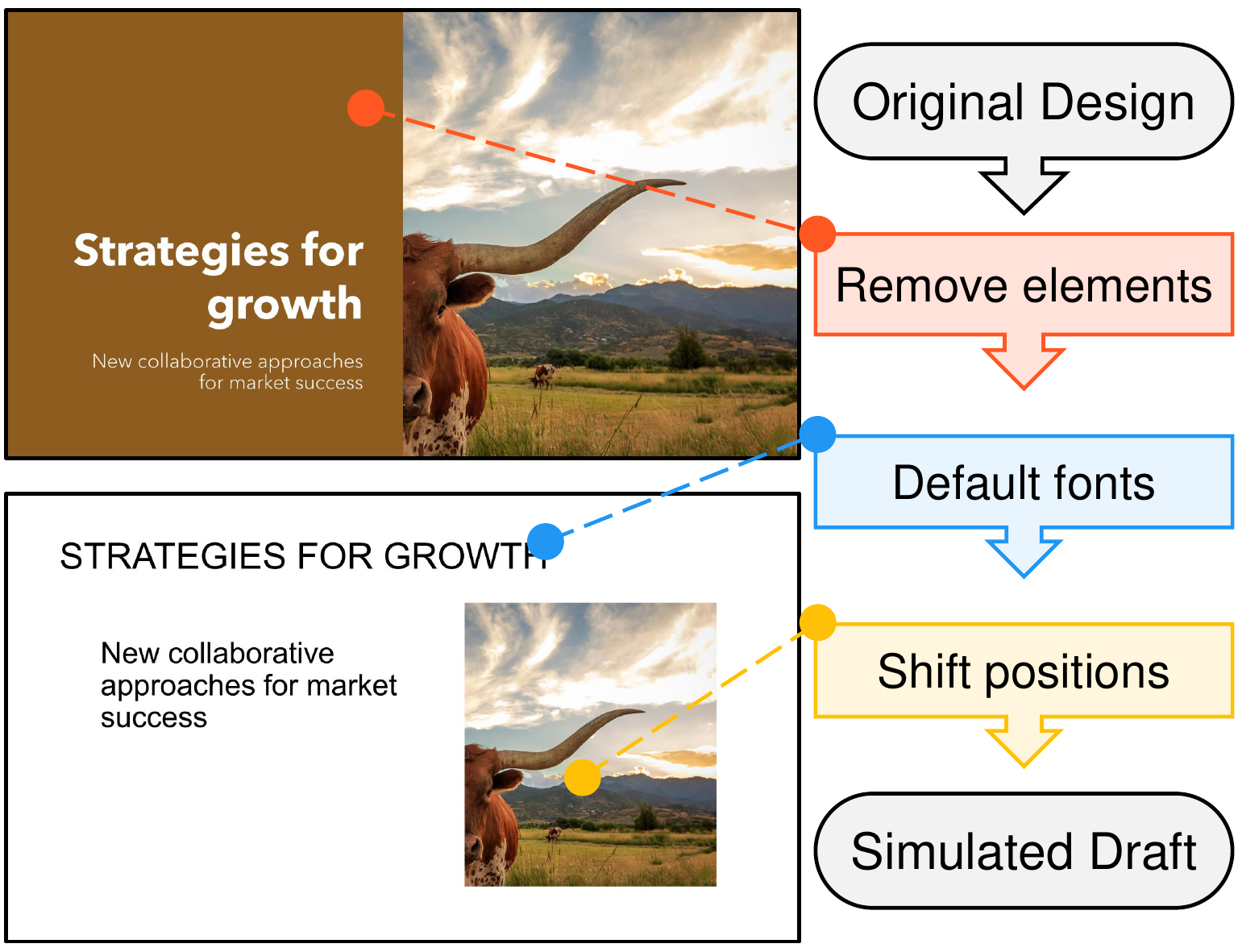}
    \vspace{-0.2cm}
    \caption{Initial drafts are simulated by removing design elements from presentation slides. }
    \label{fig:method-perturb}
    \vspace{-0.6cm}
\end{figure}

\begin{figure*}
    \centering
    \includegraphics[width=\linewidth]{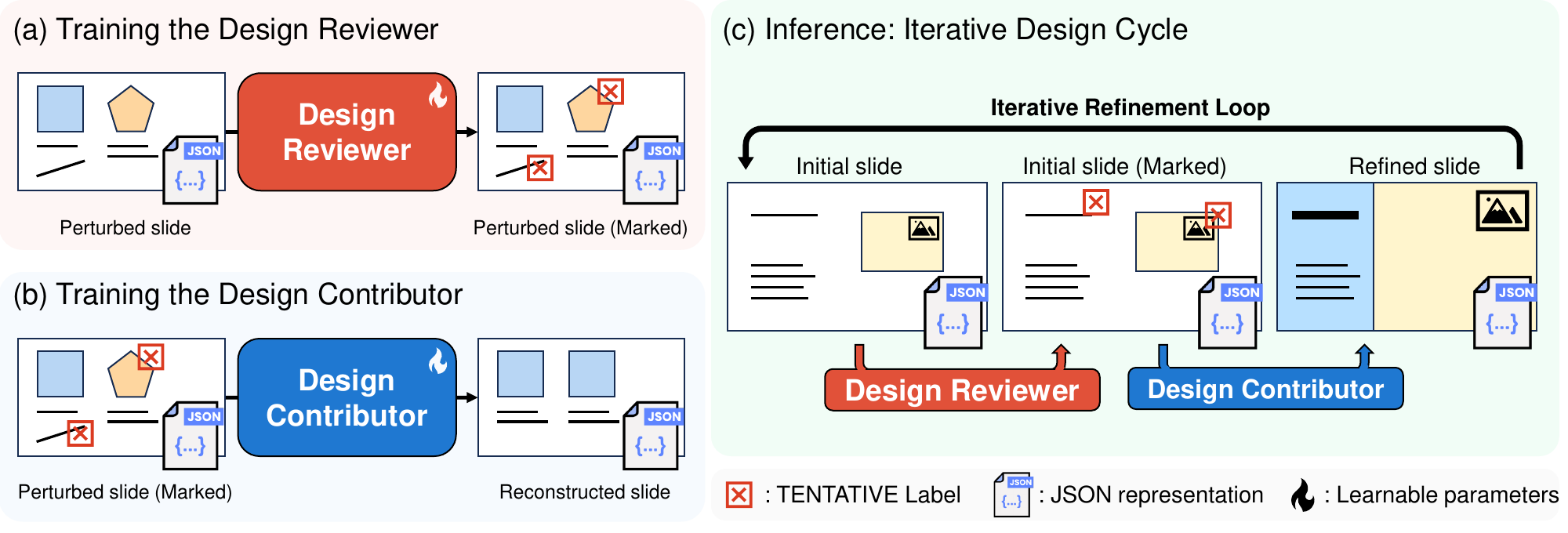}
    \caption{Overview of the DesignLab training and inference pipeline. (a) The design reviewer, an LLM fine-tuned to detect and label perturbed slide elements as TENTATIVE. (b) The design contributor, a separate model, refines the elements labeled TENTATIVE. (c) The iterative refinement process alternates between the reviewer and contributor until no elements are labeled TENTATIVE.}
    \label{fig:method-main}
    \vspace{-0.5cm}
\end{figure*}

\section{Related Work}
\paragraph{Design generation}
Generating designed documents has recently gathered attention, with approaches evolving to address a wide range of media formats ranging from document layouts~\cite{layouttransformer, layoutdm, ralf} to posters~\cite{cole, opencole}, web pages~\cite{webrpg}, and presentation slides~\cite{d2s,doc2ppt,ppsgen}.
Early work primarily focused on the placement of elements(\eg, text boxes and images)~\cite{layoutdm, ralf}, while recent methods have broadened their scope to include text attributes, such as fonts and colors~\cite{webrpg, canvasvae}, and even the creation of decorative images~\cite{cole, opencole, pptagent}.
Although these approaches now often produce designs in editable formats such as HTML or JSON, they essentially lack mechanisms to assist users to further refine the output or automatically correct errors. 

\vspace{-0.3cm}
\paragraph{Design editing}
Another line of work focuses on replacing specific elements in a completed design, such as images or text, based on manual user selection~\cite{flexdm,markupdm}. 
A recent study~\cite{autopresent} proposes an automated approach for iteratively refining slides by leveraging GPT-4o's~\cite{gpt4o} coding capabilities to adjust slide-generating scripts. 
However, in many cases, users must either identify errors themselves or rely on a single-stage refinement procedure that attempts to handle everything at once. 
In this paper, we aim to repair design flaws by addressing them in incremental steps, which is critical for handling the complexity of real-world designs. 

\section{Method}
\subsection{Design scope for presentation slides}
We focus on generating and refining the \emph{design-related elements} that are typically addressed in a real-world design process. 
These elements, detailed in \Cref{tab:method-attributes} and \Cref{sec:appx-json-scope}, include basic pre-defined shapes (\eg, rectangles, rounded rectangles, and circles), text attributes (\eg, font type, font size, and line spacing), and shape properties (\eg, position and color). 
It is important to note that we do not include generating contents itself, such as generating new text and images, within the design scope, as these are generally fixed by the user. 
Our primary goal is to enable models to identify and refine the combinations of these design attributes.

\subsection{JSON representation of presentation slides}
Presentation slides, typically stored as \verb|.pptx| files, consist of multiple XML documents~\cite{xml} that define the content of the slides. 
Due to the simplicity of XML as a markup language, converting these files into a structured JSON format is a straightforward process~\cite{autopresent}. 
Thus, we represent the design-related elements described earlier in JSON, which can be easily processed by LLMs as both inputs and outputs. 
However, we exclude media content encodings (\eg, images and videos) as they consume excessive sequence length. 
Instead, we retain only the shape and position attributes of media elements.

\begin{table}
\centering
\begin{tabular}{@{}ccc@{}}
\toprule
\multicolumn{1}{c}{Category} & Type        & Examples                 \\ \midrule
\multirow{2}{*}{Shapes}       & Auto shape  & Rectangle, Line, Circle \\
                              & Placeholder & Image, Video            \\ \midrule
\multirow{4}{*}{Attributes}   & Position    & Width, Height           \\
                              & Text        & Font Size, Font type    \\
                              & Color       & RGB values              \\
                              & Fill        & Solid, Gradient, Pattern\\ \bottomrule
\end{tabular}
\vspace{-0.2cm}
\caption{Design elements included in our design scope. }
\vspace{-0.7cm}
\label{tab:method-attributes}
\end{table}

\subsection{Defining rough drafts}
Since presentation slides are typically only available in their final form, it is difficult to obtain matched pairs of rough drafts and their corresponding refined versions. 
To resolve this issue, we simulate rough drafts by introducing random perturbations into the JSON representation of each slide. 
Specifically, we randomly remove graphical elements, shift positions, modify colors, and change font attributes, as illustrated in \Cref{fig:method-perturb} and \Cref{sec:appx-perturbations}. 
By controlling the severity of the perturbations, we can simulate different stages of the design process, ranging from near-finished slides to those that need substantial edits. 
This approach yields training pairs (perturbed drafts and their corresponding final versions), providing a framework in which our models learn to handle iterative improvement. 

\subsection{Training the design contributor and reviewer}
\paragraph{Design reviewer}
Given a pair consisting of the final design and its perturbed version, we train the design reviewer to detect which elements have been altered. 
Specifically, the design reviewer operates on the JSON representation of the perturbed design and labels any JSON elements that require improvement. 
We fine-tune a LLM for this task, as it offers a flexible and scalable solution than building a dedicated classifier, adapting to various error types with minimal architectural changes. 
As illustrated in \Cref{fig:method-main} (a), any element that requires modification is assigned a TENTATIVE status tag. 
In other words, the input of the reviewer is the JSON string of the perturbed design, while the output is the same JSON with certain elements marked as TENTATIVE. 

\paragraph{Design contributor}
While the design reviewer identifies elements that require improvement, the design contributor is trained to recover the perturbed elements, restoring them to their original design (\eg, exact positions and correct colors) based on the JSON string with TENTATIVE labels. 
The TENTATIVE labels serve as indicators that specify which elements should be changed, as the model is trained to modify only those marked as TENTATIVE. 
Importantly, the contributor is not restricted to altering existing elements, as it can also generate new ones when perturbations involve removing content. 
Through this process, the design contributor completes the iterative refinement loop by fully addressing flagged flaws and returning the slides to a high-quality final state.

\subsection{Iterative refinement}
Having trained the design reviewer and design contributor, we now bring them together in an iterative inference pipeline. 
First, all elements in the slide are initially labeled as TENTATIVE to prompt the design contributor to apply corrections or restorations. 
The contributor’s output is then returned to the design reviewer, which inspects the updated design and labels any remaining or newly introduced issues as TENTATIVE. 
This updated design is passed again to the contributor for another round of fixes, and the cycle continues until either the reviewer identifies no further issues (triggering an early stop) or a maximum iteration limit is reached. 
Through this repeated loop of targeted detection and correction, the final design emerges from multiple incremental improvements, \emph{achieving a level of polish that single-step methods typically cannot match}.

\section{Experiments}
\subsection{Dataset}
Presentation slides are widely available on various online communities and marketplaces. 
In this study, we utilize an internal dataset of 200,163 slides collected from the web.
A key reason for collecting native \verb|.pptx| or similarly structured files is their element-level decomposability. 
Unlike image renderings of designs, these vectorized formats allow us to identify and manipulate the components (\eg, text boxes and shapes).
Collecting such data has been a growing trend in design research~\cite{canvasvae, webrpg}, which is crucial for training models to generate and revise individual elements. 

While this large collection is used for training, it does not include intermediate or draft versions of slides. 
Since genuine rough drafts are rarely shared or archived publicly, we manually created a set of 77 rough drafts for evaluation
This smaller dataset includes slides with typical early-stage imperfections (\eg, misaligned elements, default font and colors), closely reflecting the type of real-world drafts we aim to improve. 
We will make these manually created rough drafts available, enabling the community to benchmark iterative refinement methods on a consistent, realistic test set.

\begin{table}[t]
\centering
\setlength{\tabcolsep}{4.5pt} 
\begin{tabular}{@{}lccx{1.2cm}@{}}
\toprule
Methods                              & Iteration & Diversity & Stability \\ \midrule
WebRPG~\cite{webrpg}                 & \RedX & \GreenCheck & \GreenCheck \\
AutoPresent~\cite{autopresent}       &  $\sim$   & \RedX & \RedX \\
Powerpoint Designer~\cite{msdesigner}& \RedX & \RedX & \GreenCheck \\ \midrule
Ours                                & \GreenCheck & \GreenCheck & \GreenCheck \\ \bottomrule
\end{tabular}
\vspace{-0.2cm}
\caption{Comparison of baseline approaches and our method across key traits ($\sim$ indicates partial support).}
\vspace{-0.5cm}
\label{tab:exp-baselines}
\end{table}

\subsection{Baselines}

\paragraph{WebRPG}
To evaluate the benefits of our iterative design approach, we compare against a single-step baseline that attempts to improve a perturbed slide in one step. 
Specifically, using the same dataset, we fine-tune a LLM to generate an enhanced design directly from the perturbed input. 
In spirit, this single-step method parallels WebRPG~\cite{webrpg}, which similarly focuses on producing an updated layout in one go, but for websites rather than presentation slides. 

\vspace{-0.4cm}
\paragraph{AutoPresent}
AutoPresent~\cite{autopresent} is a recent agent-based system~\cite{voyager,toolformer,react,ttys} designed to generate new slides from scratch based on instructions or a reference slide. 
The system leverages the coding capabilities of GPT-4 to refine slides by producing Python scripts that modify the original slides based on few-shot exemplars. 
Instead of directly generating an improved layout, AutoPresent generates these refining scripts. 
While this approach technically supports iterative refinement, the high failure rate makes it impractical for repeated use. 

\vspace{-0.4cm}
\paragraph{Powerpoint Designer}
Lastly, we compare our method against PowerPoint Designer~\cite{msdesigner}, a commercial feature within Microsoft PowerPoint. 
The Designer function attempts to fit a presentation slide into a selection of pre-defined templates, offering multiple suggestions. 
When a suitable template is found, the Designer automatically reformats to align text, images, and other elements. 
However, if no template matches the structure of the content, the Designer provides no suggestions at all. 

We summarize the key characteristics and capabilities of these baselines in \Cref{tab:exp-baselines}.

\subsection{Implementation details}
We fine-tune two instruction-tuned Qwen2.5-1.5B models~\cite{qwen2, qwen2.5} for the reviewer and contributor roles separately. 
The models are trained on our dataset for 400,000 steps with a learning rate of 1e-4, utilizing a warm-up phase and the AdamW optimizer~\cite{adamw}. 
The models can be run with less than 8GB of VRAM, making them suitable for deployment on commercial GPUs. 
The chat templates used for training are provided in Appendix~\Cref{sec:appx-training}.

\begin{figure*}
    \centering
    \includegraphics[width=\linewidth]{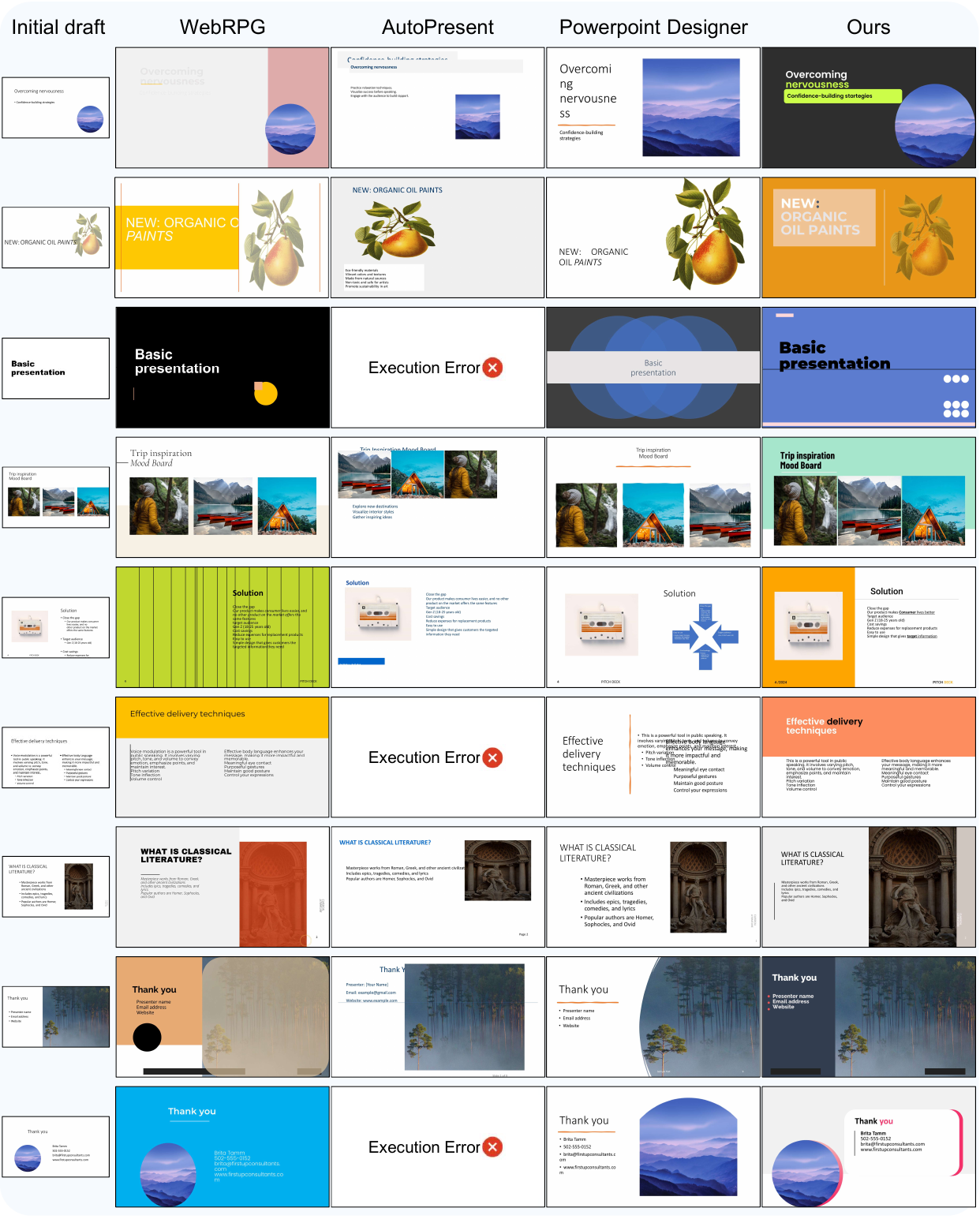}
    \vspace{-0.5cm}
    \caption{Qualitative comparison of slide refinement results on manually created initial drafts. Best viewed digitally. }
    \vspace{-0.5cm}
    \label{fig:exp-quali}
\end{figure*}

\begin{figure}
    \centering
    \includegraphics[width=0.95\linewidth]{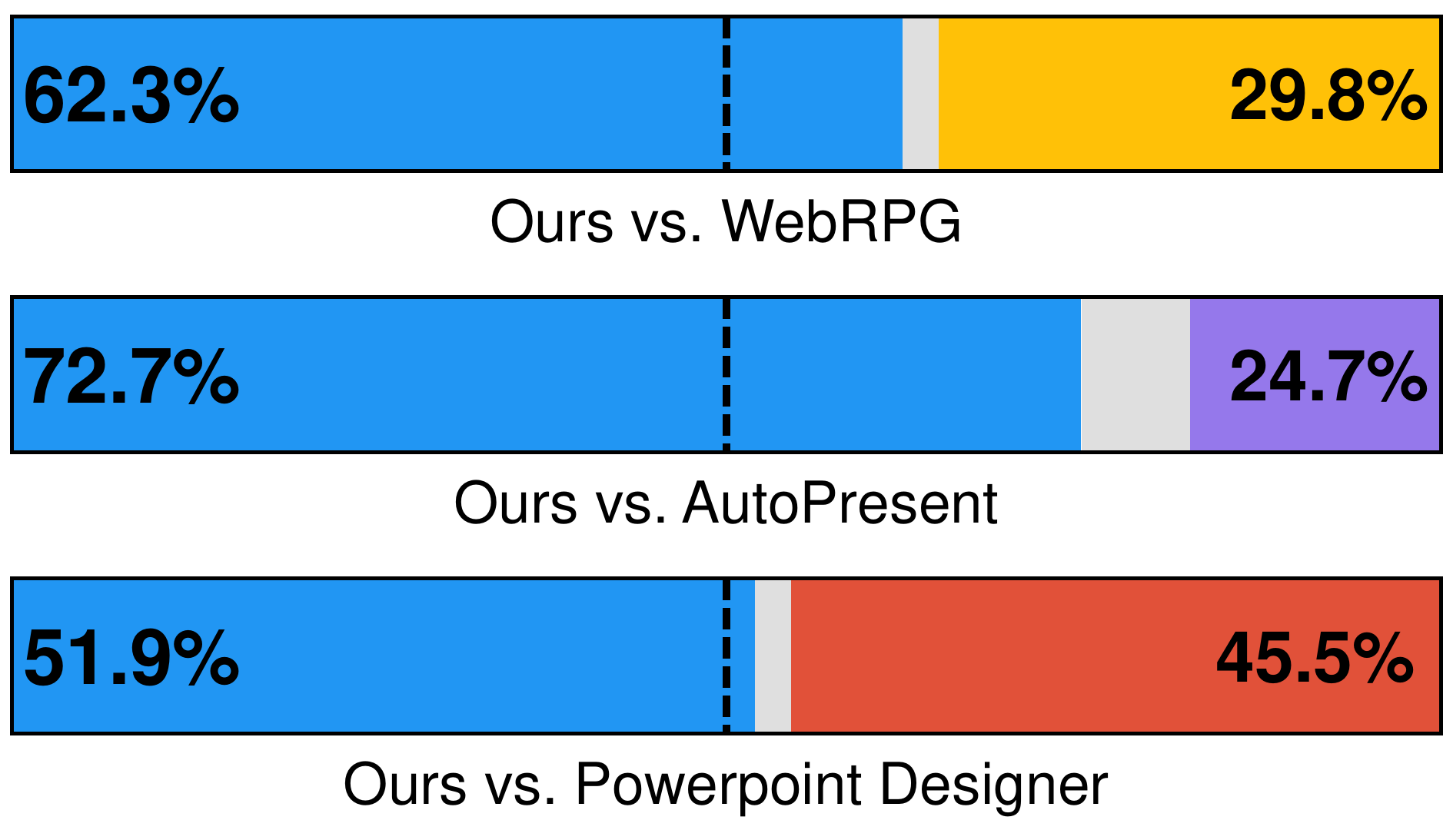}
    \vspace{-0.3cm}
    \caption{GPT-4o preference of refined slides. Left bars represent preference for our slides, with gray indicating a tie; right bars show preference for baselines. Best viewed in color. }
    \label{fig:exp-chatgpt}
\end{figure}

\subsection{Qualitative assessment}

In \Cref{fig:exp-quali}, we provide extensive qualitative comparisons of the refined presentation slides produced by each baseline approach. 
Our iterative design process yields high-quality designs compared to existing design refinement approaches, which often produces suboptimal designs. 
For example, template-based approaches, such as PowerPoint Designer~\cite{msdesigner}, fail to make refinements when no suitable template is found and lack design diversity. 
LLM-based methods, including both fine-tuned (WebRPG) and non-fine-tuned (AutoPresent) models, generate incomplete designs that require further user input to be suitable for final presentations. 

One issue we observed with AutoPresent~\cite{autopresent} is the unreliability of its scripts, which often fail to execute correctly. 
This undermines the iterative refinement process, as repeated rounds of refinement increase the likelihood of execution failures, ultimately reducing the system's effectiveness. 
Other methods also lack support for iterative refinement, as they do not accept partially designed drafts as inputs. 
In contrast, our approach ensures reliable design generation and supports progressive refinement, as our models are trained to detect and correct designs at any level of completeness.

\subsection{Quantitative comparison and assessment}

\paragraph{Design aesthetic comparison}
To quantify the difference between designs, we evaluate the designs using GPT-4o~\cite{gpt4o} similar to techniques used in recent studies~\cite{pptagent, autopresent}. 
Unlike previous approaches that evaluate designs based solely on aesthetic scores, we present two design candidates (each from a different model) and ask which one better improves the initial draft. 
This comparative approach provides a more direct assessment of each model's ability to enhance slide quality. 
As shown in \Cref{fig:exp-chatgpt}, our method consistently outperforms existing design refinement approaches, including a \emph{dedicated commercial tool}, Powerpoint Designer.
Both quantitative and qualitative results demonstrate the effectiveness and reliability of our iterative refinement process in producing high-quality, polished designs.

\begin{figure}
    \centering
    \includegraphics[width=0.95\linewidth]{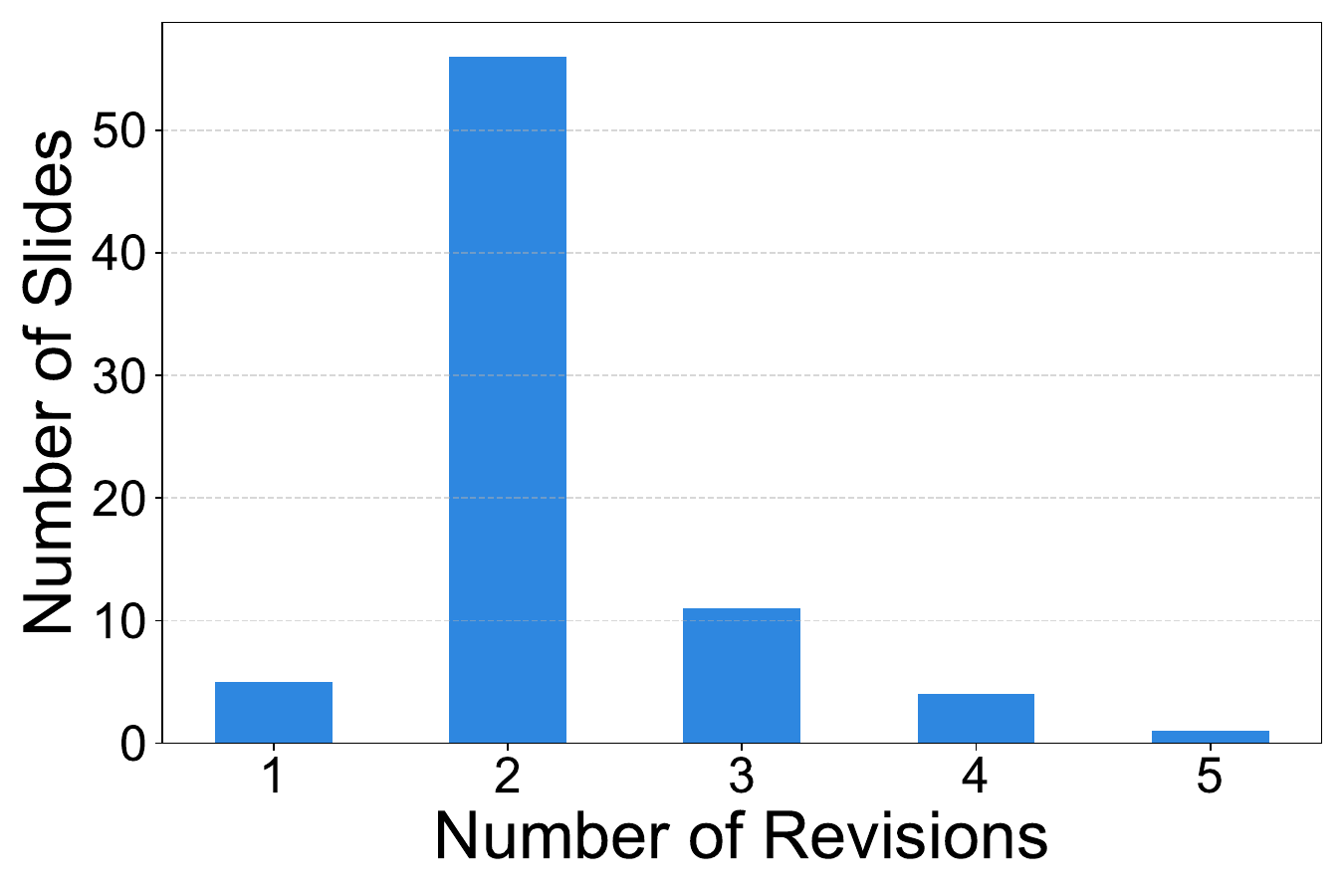}
    \vspace{-0.5cm}
    \caption{Distribution of slides across the number of iterations required to converge. }
    \label{fig:analy-iter}
\end{figure}
\begin{figure*}[t]
    \centering
    \includegraphics[width=\linewidth]{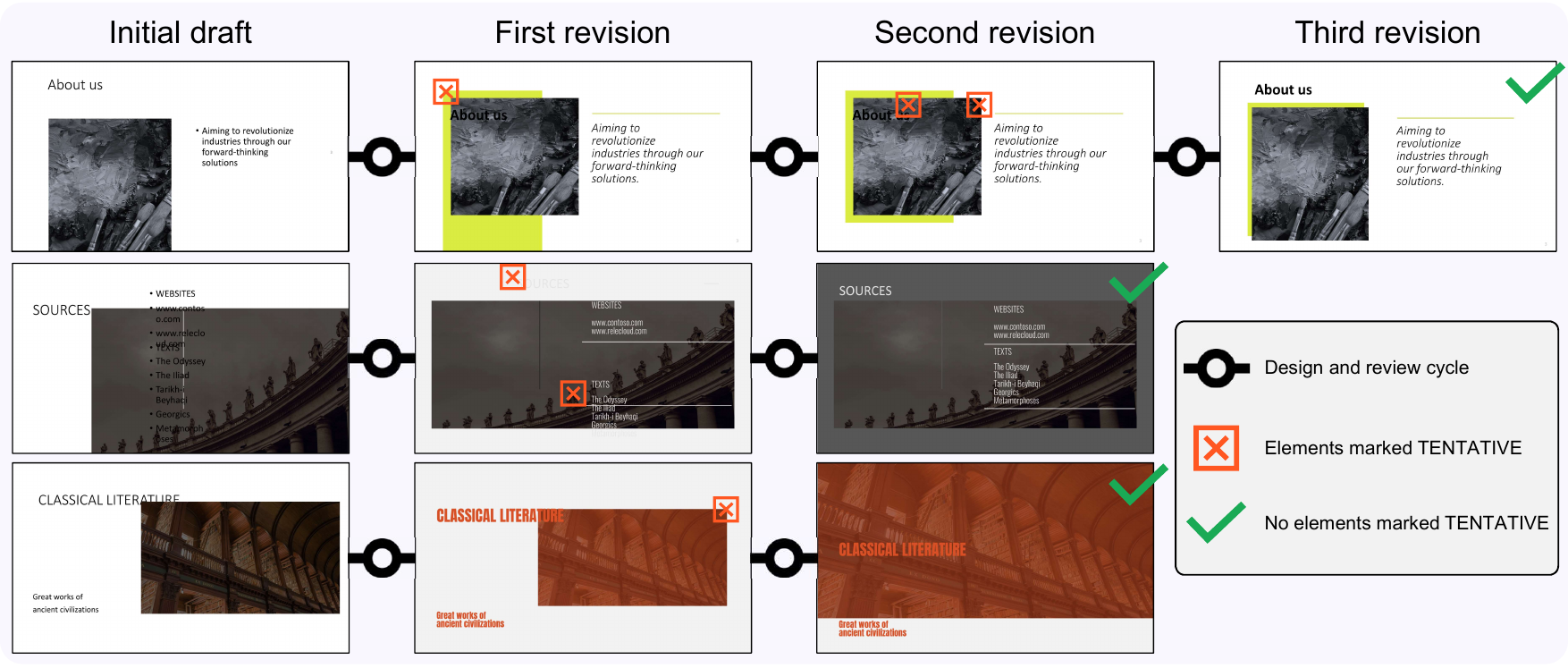}
    \vspace{-0.5cm}
    \caption{Step-by-step examples of the iterative refinement process. Each cycle of reviewing and improves the design while revealing new flaws to be revised. The cycle halts when there is no more errors are detected. }
    \label{fig:exp-iter}
\end{figure*}

\paragraph{Convergence of the iterative design cycle. }
We provide examples of the iterative refinement cycle in \Cref{fig:exp-iter}, which terminates when the design reviewer labels no more elements as TENTATIVE. 
We plot the number of revisions it takes before the design cycle come to an end in \Cref{fig:analy-iter}. 
Most slides require more than one design cycle to converge, emphasizing the need for continuous and progressive improvements. While the majority of designs converge within two iterations, the need for additional rounds highlights the complexity of the design challenges. 
This underscores the value of our approach, which facilitates ongoing refinement through multiple cycles of feedback, closely mirroring the iterative nature of real-world design processes.

\begin{figure}
    \centering
    \includegraphics[width=\linewidth]{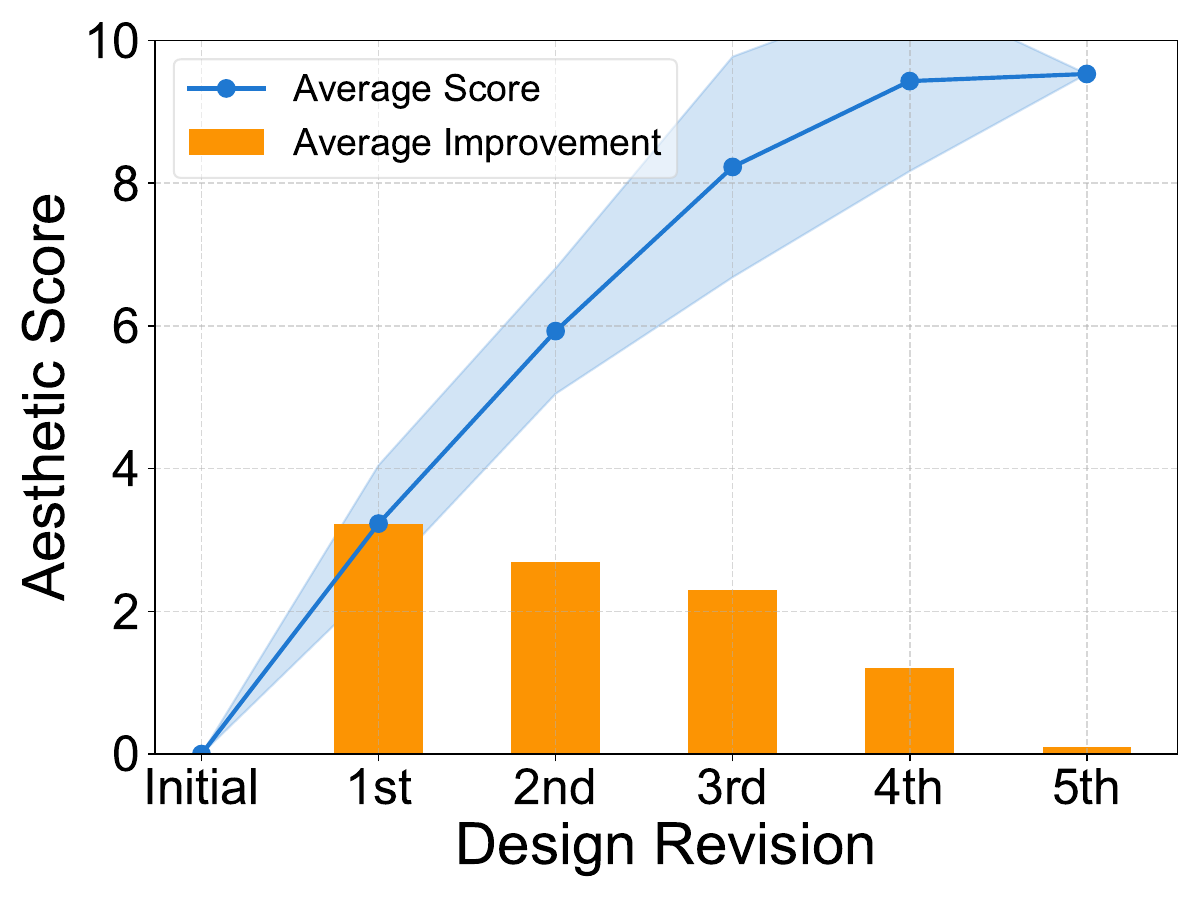}
    \vspace{-0.5cm}
    \caption{User study evaluating the effect of iterative refinement on slide aesthetics, with the shaded area representing the range within one standard deviation.}
    \label{fig:analy-improv}
    \vspace{-0.3cm}
\end{figure}

\vspace{-0.2cm}
\paragraph{Evaluating the impact of iterative refinements. }
For a quantitative analysis on the effectiveness of the iterative refinement process, we conduct a user study to assess the improvements made at each revision. 
Specifically, we asked 32 users to rate the aesthetic quality of slides on a scale of 1 to 10, given a pair of 45 slides before and after a revision (total of 90 slides). 
The distribution of scores are plotted in \Cref{fig:analy-improv}, which shows a clear trend of increasing aesthetic scores after each iteration, indicating that users perceive a clear improvement.
This progressive increase in ratings reflects the efficacy of the iterative process in addressing design flaws. 

Furthermore, the scores eventually converge, suggesting that the iterative cycle reaches a point of diminishing returns, where further revisions yield little additional perceptible improvement. 
This supports the notion that multiple rounds of feedback lead to a substantial enhancement in the overall quality of the presentation slides. 


\section{Analysis}
\subsection{Reviewer and contributor performance}
A key requirement that facilitates our iterative framework is the high accuracy of the design reviewer in detecting errors and the high responsiveness of the design contributor in correcting the identified elements. 
A highly accurate reviewer is crucial, as it ensures that design flaws are correctly identified, distinguishing between elements that should be retained and those that need to be changed. 
Similarly, a responsive design contributor is vital, as it ensures that identified errors are promptly corrected, allowing remaining issues to be detected and addressed in subsequent iterations.

To evaluate this, we measure the reviewer's performance in detecting simulated perturbations in terms of precision and recall, as well as the design contributor's responsiveness in making the necessary corrections. 
The precision and recall is measured by detecting whether randomly perturbed elements are properly labeled as TENTATIVE by the design reviewer. 
We report the precision and recall for each type of perturbation (\ie, shifted placement, duplicate shapes, altered colors, and text attributes) in \Cref{tab:analysis-accuracy}. 
The responsiveness of the design contributor is measured by verifying whether an element that has been labeled TENTATIVE has been altered by the design contributor. 
We also measure the responsiveness across different types of perturbations.

\begin{table}[t]
\centering
\begin{tabular}{@{}ccc|c@{}}
\toprule
\multirow{2}{*}{Design flaws} & \multicolumn{2}{c|}{Reviewer} & Contributor    \\ \cmidrule(l){2-4} 
                              & Precision       & Recall      & Responsiveness \\ \midrule
Shape Placement               & 0.769           & 0.149       & 1.000          \\
Shape Removal                 & 0.739           & 0.657       & 1.000          \\
Color Attributes              & 0.856           & 0.721       & 0.986          \\
Text Attributes               & 0.871           & 0.730       & 0.957          \\ \bottomrule
\end{tabular}
\caption{Accuracy of the reviewer and the responsiveness of the contributor, measured on the evaluation set.}
\label{tab:analysis-accuracy}
\vspace{-0.3cm}
\end{table}

\begin{figure*}
    \centering
    \includegraphics[width=\linewidth]{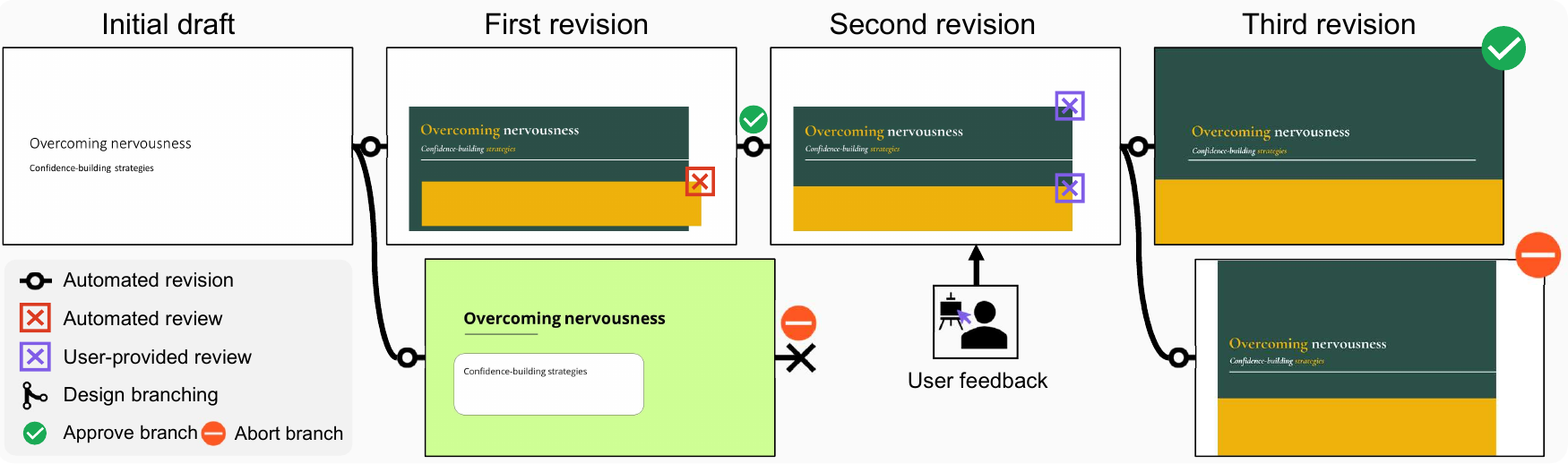}
    \caption{Interactive scenario of our design cycle featuring a branching strategy. In this scenario, the user takes the role of a design reviewer, where they can 1) choose the preferred design from two candidates, and 2) select specific elements for modification to enhance the design further. Best viewed in color. }
    \label{fig:analysis-interactive}
    \vspace{-0.3cm}
\end{figure*}

The overall precision of the reviewer is high, demonstrating that the reviewer is reliable in practice. 
Meanwhile, the design contributor demonstrates strong responsiveness when adjusting elements labeled TENTATIVE, as evidenced by the high-quality outcomes shown in \Cref{fig:exp-quali} and \Cref{fig:exp-iter}. 
We believe this high responsiveness is relatively easy to achieve because the contributor receives explicit labels identifying which elements need changes. 
However, by isolating error detection within the design reviewer, we reduce the cognitive load on the design contributor, which can then focus solely on producing the necessary corrections in the design.

\begin{figure}
    \centering
    \includegraphics[width=0.9\linewidth]{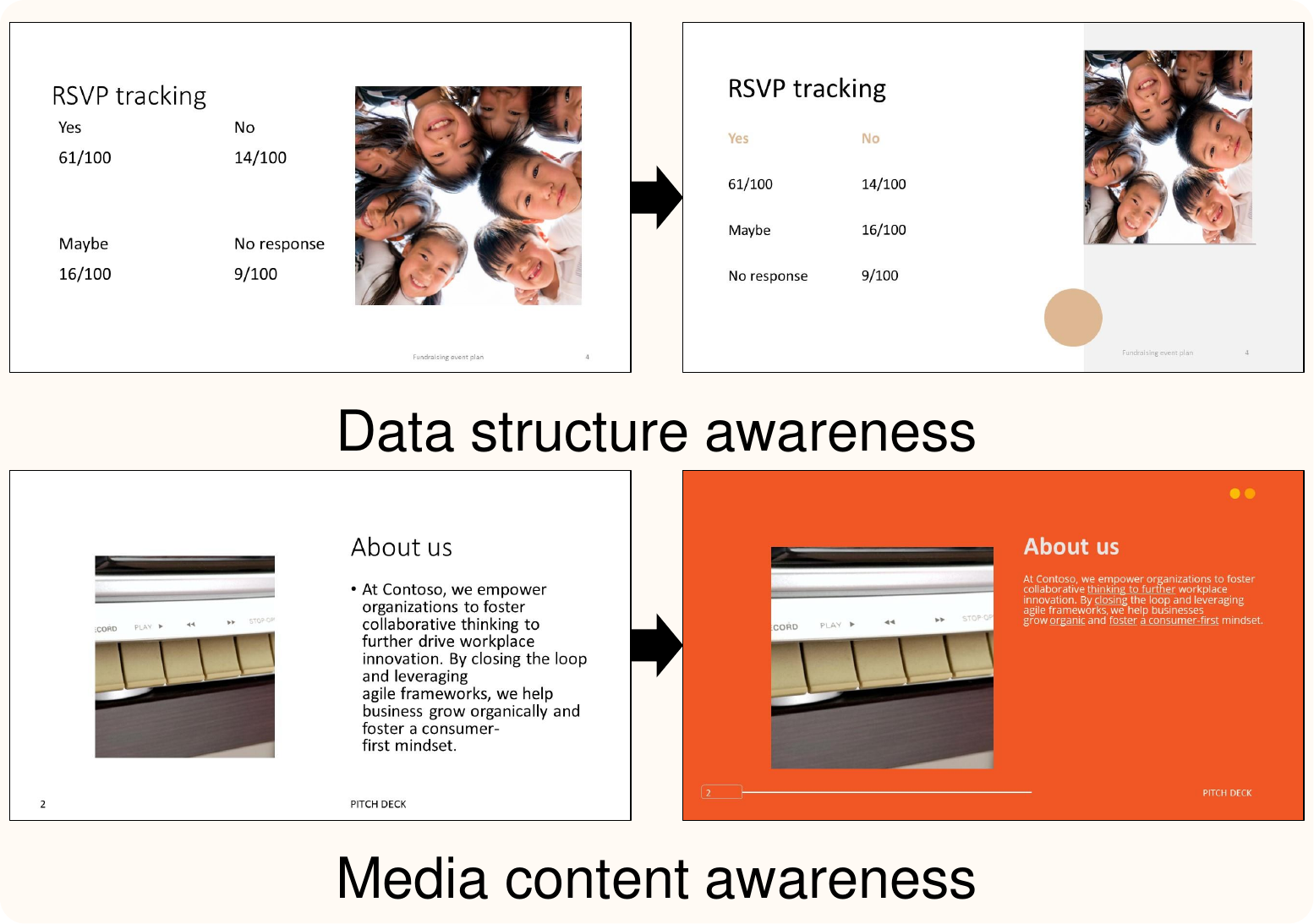}
    \vspace{-0.2cm}
    \caption{Failure cases of our approach with complex data structures and content awareness. }
    \label{fig:analysis-fail}
    \vspace{-0.4cm}
\end{figure}

\subsection{Interactive scenarios and branching strategies}
Certain design flaws, such as subtle shifts in position, are inherently difficult to detect, as reflected by the low position recall in \Cref{tab:analysis-accuracy}. 
To address these cases, our framework supports interactive use cases where auser can take on the role of reviewer, manually selecting elements that require changes. 
This approach enables further refinements customized to the user's preference.

Additionally, our system can generate multiple ``branches'' of a design, akin to branching in software development pipelines, at minimal cost via batched inference. 
Each branch represents a distinct design suggestion, allowing users to compare options and select (merge) those that best fit their preferences. 
We demonstrate an example of this interactive process in \Cref{fig:analysis-interactive}. 
By combining branching with user-driven reviewing, we offer a flexible design tool that enables users to iteratively refine their initial drafts and preserve or discard elements as needed.

\subsection{Failure cases}
We illustrate two failure cases of our approach in \Cref{fig:analysis-fail}. 
One issue we observe is that our model sometimes struggles to fully comprehend complex data structures represented in text format, such as tables and graphs. 
We aim to address these challenges in future updates by leveraging larger models (with 7B, 14B, and 32B parameters) that possess a deeper understanding of these structures.

Additionally, since our model does not encode any media content (\eg, images or videos), it is unable to interpret their visual content and their colors.
As a result, the designs it generates may feature colors that do not align with those present in the media. 
This design choice of not encoding images stems from the fact that VLMs~\cite{llava,llavanext,qwenvl} at their current stage consume too many tokens, and JSON representations already occupy a significant portion of the token budget. 
We plan to make future updates by incorporating meta-information, such as content tags and color palettes. 

\section{Conclusion}

In this paper, we present a design assistant tool that can iteratively detect and revise design flaws, modeling real-world designing workflows. 
Unlike previous approaches, we focus on revising imperfect drafts, a critical yet often overlooked aspect in studies.
Our experiments, including a user study, demonstrate that our iterative process of revising intermediate draft produces high-quality outputs, consistently improving the design over time. 
We believe our approach has broader implications beyond presentation slides and offers a general framework for making design tasks more accessible and efficient across diverse design domains. 

{
    \small
    \bibliographystyle{ieeenat_fullname}
    \bibliography{main}
}

\clearpage
\setcounter{section}{0}
\setcounter{page}{1}
\renewcommand*{\thesection}{\Alph{section}}
\maketitlesupplementary

\section{Extended Related Work}
We further organize the related work section as, our work sits at the intersection of several research areas. 

\paragraph{Automated design assessment and critique generation.}
A growing body of work focuses on automatically evaluating and providing feedback on visual designs. 
UIClip~\cite{uiclip} introduces a data-driven approach for assessing user interface designs, demonstrating that machine learning models can learn to evaluate design quality across multiple dimensions such as aesthetics, usability, and accessibility. Similarly, \citeauthor{duan2024generating} explore generating automatic feedback on UI mockups using large language models, showing that LLMs can provide constructive design critiques that help designers identify improvement opportunities. These works establish the feasibility of the ``reviewer'' role in our framework, though they focus primarily on evaluation rather than the iterative refinement process we propose.

\paragraph{Visual design beyond presentation slides.}
While our work focuses on presentation slides, it connects to broader research in automated visual design generation and assessment. 
User interface design has been a particularly active area, with researchers developing systems for generating, evaluating, and refining interface layouts~\cite{duan2024generating}. 
Poster design~\cite{creatiposter,posta}, web layout optimization~\cite{webrpg}, and graphic design automation share similar challenges around balancing aesthetic principles with functional requirements. 
Our JSON-based representation approach and perturbation-based training methodology could potentially extend to these domains, as they face similar challenges in capturing design elements and learning improvement patterns.

\paragraph{Synthetic data generation for design understanding.}
A key challenge in training design systems is the scarcity of paired examples showing design evolution from rough drafts to polished versions. 
DreamStruct~\cite{dreamstruct} addresses this challenge by generating synthetic data to understand slides and user interfaces, demonstrating that artificial data creation can effectively support design-related machine learning tasks. 
Their work validates our approach of using synthetic perturbations to simulate rough drafts, though our focus on creating iterative refinement pairs differs from their broader data generation objectives. 
This synthetic data approach has proven valuable across various design domains where naturally occurring before-and-after examples are rare.

\begin{figure}
    \centering
    \includegraphics[width=\linewidth]{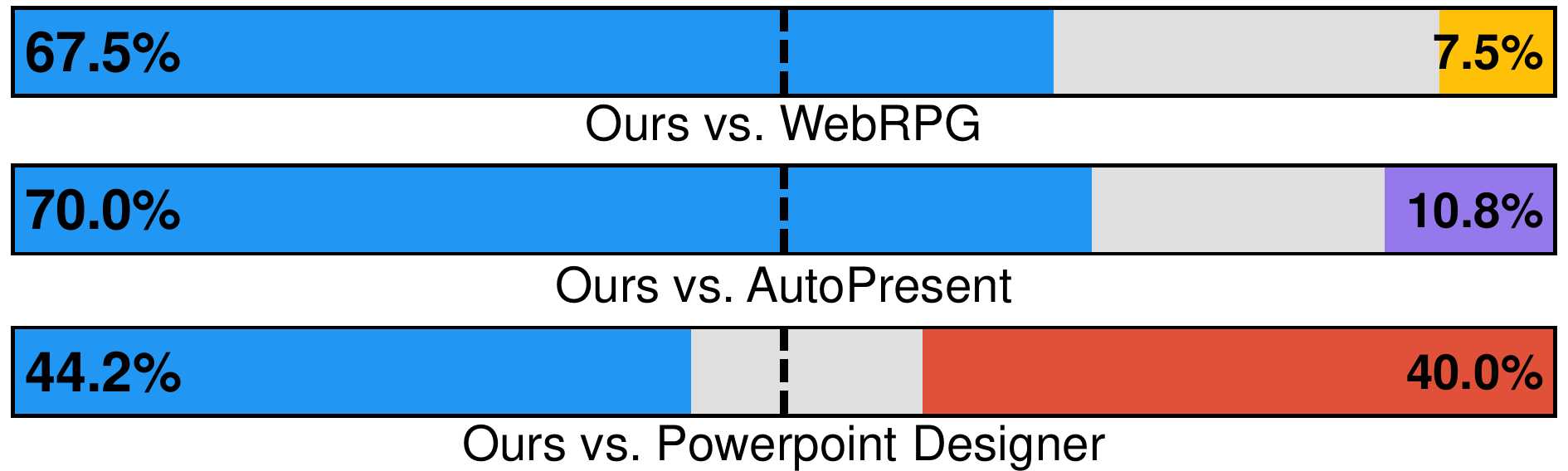}
    \caption{User preference of slides refined by our approach and each baseline approach. }
    \label{fig:appx-user-study}
    \vspace{-0.5cm}
\end{figure}

\section{User Study on Refined Slides}
To validate our evaluation methodology, we conducted a user study involving 20 participants. 
This study was designed to parallel our GPT-4o evaluation, with participants assessing the same presentation slides using identical criteria and rating scales. 
The human evaluation results are presented in \Cref{fig:appx-user-study} and demonstrate patterns remarkably consistent with our GPT-4o assessments, supporting the reliability of our automated evaluation approach.

\section{JSON Representation of Presentation Slides}

\subsection{What content do we represent?}
\label{sec:appx-json-scope}
Presentation slides, typically in \texttt{.pptx} format, consist of multiple XML documents~\cite{xml} that describe each slide's structure.
Since XML is a well-documented markup language, its interpretation is relatively straightforward.
Open-source libraries, such as \texttt{python-pptx}, provide robust support for parsing these formats.
However, XML files are often lengthy and not well-suited as direct inputs or outputs for LLMs.
To address this, we transform each XML file in our dataset into a structured JSON format that conveys the slide's content in natural language.

During this process, we selectively extract relevant slide information.
We categorize design elements, as shown in \Cref{tab:appx-shapes}, into shapes and attributes.
Shapes are the visible entities in a slide, including pre-defined basic shapes (\eg, circles, ovals, and rounded rectangles) and placeholders for media content such as images and videos.
We consider 34 basic shape types supported in PowerPoint but exclude complex elements like tables, graphs, and plots for simplicity.
Attributes define shape properties, including position, text content, colors, and fill types, and are always associated with a shape.

\subsection{Why do we use a JSON representation?}
\label{sec:appx-why-json}
It is important to note that representing slides in a structured, text-based format (JSON) offers significant flexibility. 
This approach allows us to seamlessly incorporate additional attributes such as transparency levels, gradient fills, or other visual properties without requiring major architectural changes often required by previous approaches~\cite{flexdm}. 
Consequently, our representation enables a highly expressive generation process capable of handling a wide variety of design elements, making it adaptable to diverse design requirements.

Ultimately, the expressive nature of our JSON representation defines the scope of our model's generation capabilities, allowing the model to manipulate and refine multiple aspects of slide design comprehensively. 
By clearly delineating the types of elements and attributes included in our representation, we also precisely define the scope and boundaries of the design improvements our model is capable of generating. 

Vision-language models (VLMs) present an interesting future direction for visually inspecting slide components, but image-based approaches remain outside the scope of our current work due to token efficiency considerations. 
Our JSON-based implementation proves effective for core design tasks like text box alignment and color suggestions, showing that structured representations can successfully enable iterative design improvements.

\begin{table}[t]
\begin{tabular}{@{}ccc@{}}
\toprule
Category & Type         & Examples                 \\ \midrule
\multirow{2}{*}{Shapes}     & Autoshapes & \begin{tabular}[c]{@{}c@{}}Line, Circle, Oval, \\ Rectangle, \\ Rounded Rectangle, \\ Trapezoid, Arrow\end{tabular} \\ \cmidrule(l){2-3} 
         & Placeholders & Image, Video             \\ \midrule
\multirow{4}{*}{Attributes} & Position   & \begin{tabular}[c]{@{}c@{}}X-coordinate, Y-coordinate\\ Width, Height\end{tabular}                                  \\ \cmidrule(l){2-3} 
         & Text         & \begin{tabular}[c]{@{}c@{}}Font Type, Font Size,\\ Line Width, Text Alignment\end{tabular}   \\ \cmidrule(l){2-3} 
                            & Color      & \begin{tabular}[c]{@{}c@{}}RGB values,\\ Transparancy values\end{tabular}                                           \\ \cmidrule(l){2-3} 
         & Fill         & Solid, Gradient, Pattern \\ \bottomrule
\end{tabular}
\caption{Categorization of design elements within our scope of generation.}
\label{tab:appx-shapes}
\end{table}

\section{Training Details}
\label{sec:appx-training}
\subsection{Model selection and hyperparameters}
We use an instruction-tuned Qwen2.5-1.5B model~\cite{qwen2.5} to serve as both the reviewer and the design contributor.
Although we do not train our model on other LLMs~\cite{llama3,mistral,deepseek}, we believe that these differences are marginal, as we excessively fine-tune the model to perform their roles. 

We train our model on 8 Nvidia A100 GPUs using a batch size of 1 for 400,000 steps, using a learning rate of 1e-4 with a linear warmup for 500 steps. 
We find that we need long fine-tuning steps to achieve a decent performance, largely due to learning the JSON representation of slides. 
However, we find the training process itself to be stable and is robust to different training hyperparameters. 

\subsection{Training samples for supervised fine-tuning}
\label{sec:appx-perturbations}
\begin{figure}
    \centering
    \includegraphics[width=\linewidth]{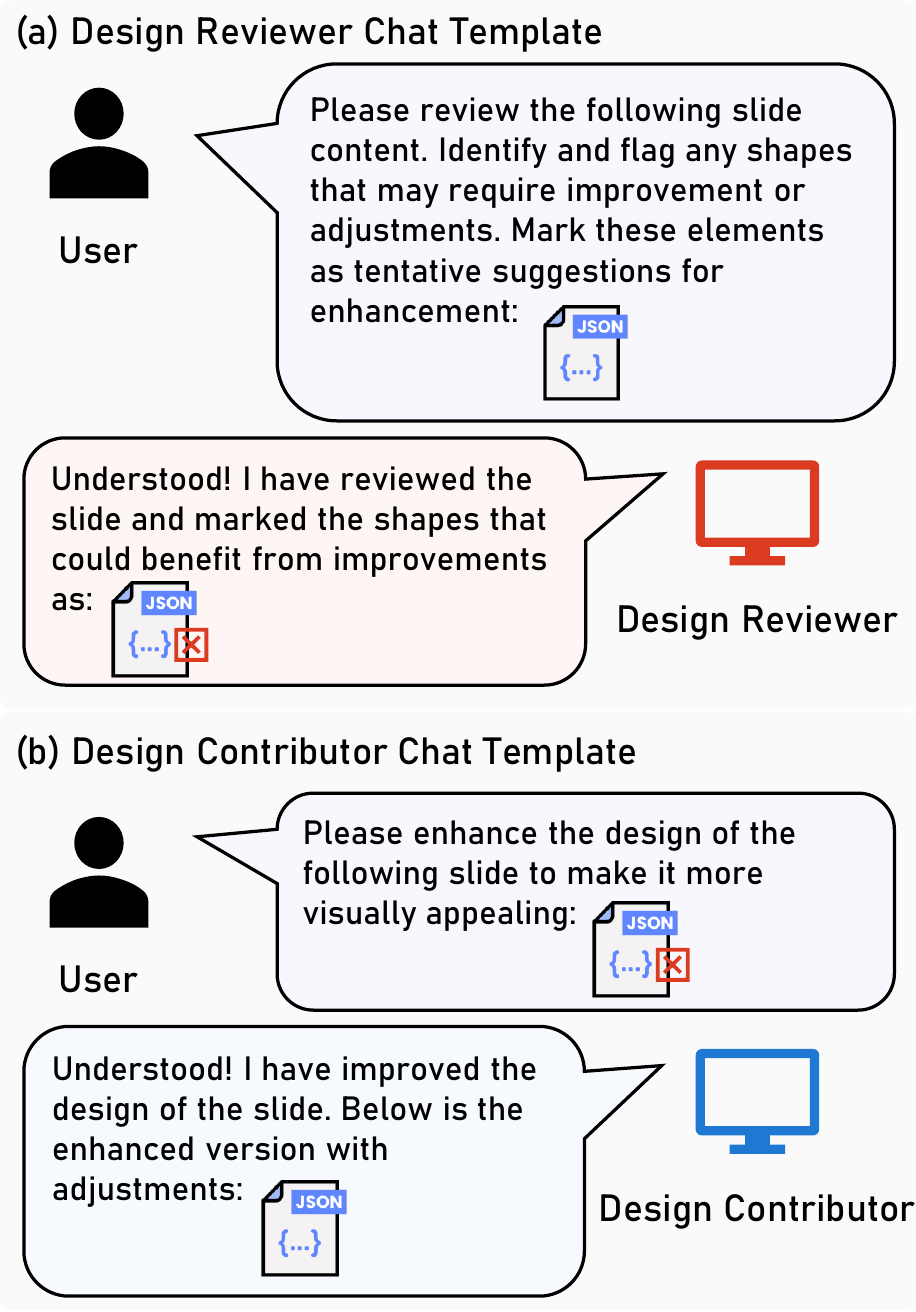}
    \caption{Chat templates used for supervised fine-tuning. The JSON representation of slides is inserted into the template. JSON icons with X marks indicate files containing TENTATIVE labels. }
    \label{fig:appx-chat-template}
\end{figure}

The reviewer identifies design flaws in a slide and labels them as TENTATIVE.
The contributor then improves the slide based on its JSON representation, incorporating the TENTATIVE labels. 
Since both roles rely on instruction-tuned models~\cite{instructiontune}, we structure our supervised training samples using a simple chat template, as shown in \Cref{fig:appx-chat-template}.
These templates are also used during inference.

We further provide details on how we perturb the slides (simulate rough drafts). 
Every shape and attributes summarized in \Cref{tab:appx-shapes} are a subject to perturbation. 
For shapes, we either remove a certain autoshapes entirely, training the model to generate shapes if needed, and also duplicate existing shapes, training the model to remove shapes if need. 
For attributes, we randomly shift the positions and alter colors for continuous values. 
We set categorical values, such as text attributes, to a set of pre-defined default values. 
For example, for font types, we change the font to one of the popular fonts (\eg, Arial, Roboto, and Calibri). 
Note that this doesn't mean that our design refinement is limited to these fonts, as the design contributor is trained to generate font names in the original slides.

\section{Inference Cost}
\label{sec:appx-inference}
Inference is performed using the same chat template used during training. 
We observe that inference is memory-efficient and consistently fits within the 2048-token maximum sequence length, despite the detailed JSON representation of the slides. 
Without any acceleration techniques~\cite{bf16, gptq, awq}, both the design reviewer and contributor complete slide generation in under 30 seconds and require 8GB of VRAM using well under typical memory limits.
Both the design reviewer and contributor generates a presentation slide under 30 seconds, without any acceleration techniques. 
Additionally, inference with an optimized implementation like vLLM~\cite{vllm} significantly reduces runtime, finishing each step within 6 seconds. 
This demonstrates that our framework is suitable for practical, real-time interactive scenarios, even on hardware with limited computational capacity.

\begin{figure}[t]
    \centering
    \includegraphics[width=\linewidth]{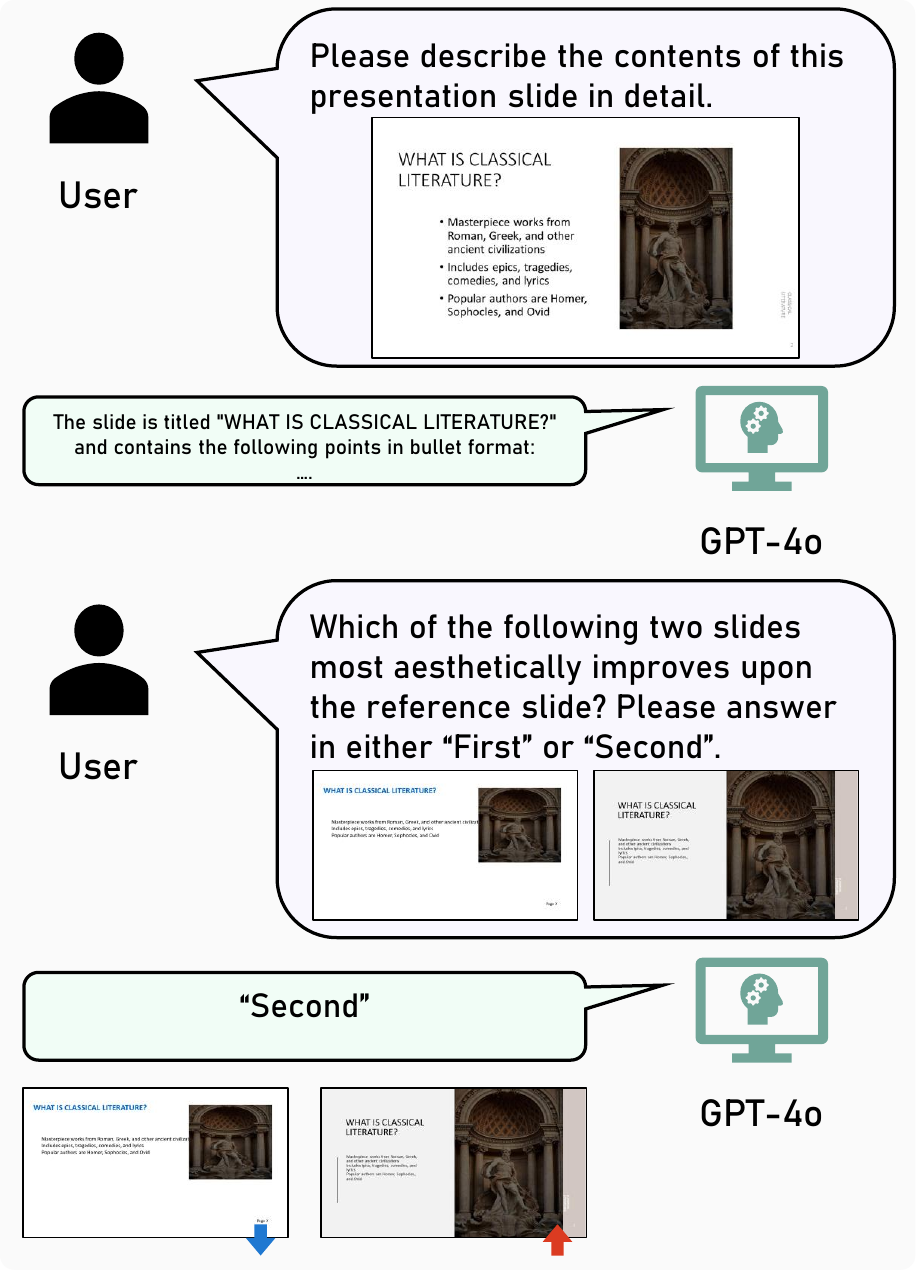}
    \caption{Comparing two versions of refined presentation slides using GPT-4o. }
    \label{fig:appx-chatgpt}
\end{figure}

\section{Evaluation Details}
\label{sec:appx-chatgpt}
To compare the aesthetic quality of slides, we prompt GPT-4o to select which of two slides represents a better refinement relative to an initial draft. 
Following recent evaluation methodologies leveraging GPT models, we employ a simple chain-of-thought reasoning strategy~\cite{cot}, encouraging GPT-4o to first explicitly analyze each slide before making a decision. 
Specifically, GPT-4o initially examines the provided rough draft, then compares two refined slides, and finally selects the slide demonstrating superior improvement. 
By repeating this procedure across multiple comparisons, we compute pairwise win rates for each method. 
We believe that this is a more reliable way to evaluate designs, compared to asking a model to rate a score for each draft, as these comparisons are easier to make for humans as well. 
The entire evaluation workflow is illustrated in \Cref{fig:appx-chatgpt}.

\section{Additional Results}

\begin{figure*}[t]
    \centering
    \includegraphics[width=\linewidth]{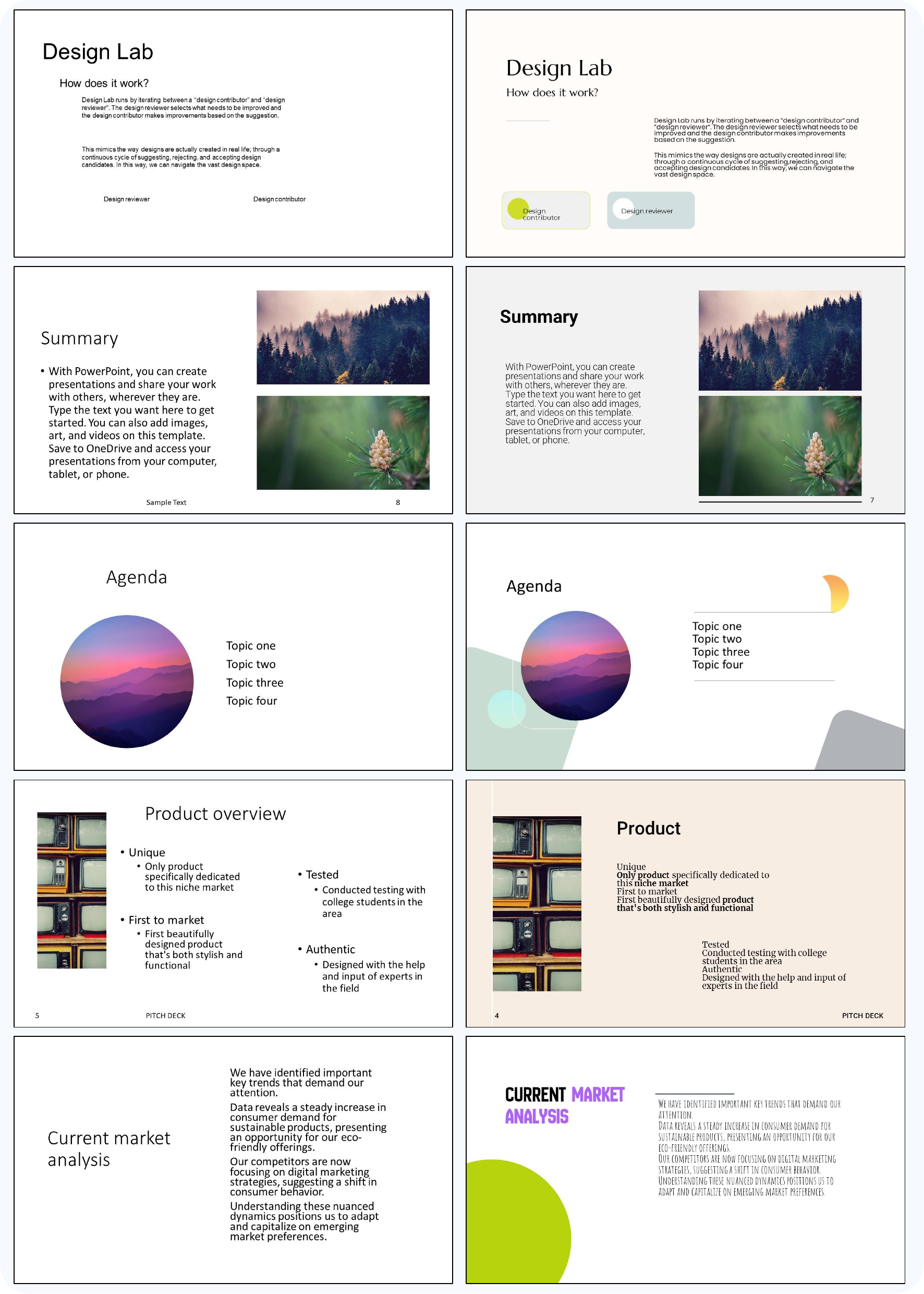}
    \label{fig:appx-qauli}
\end{figure*}

We present additional qualitative results in \Cref{fig:appx-qauli} to further illustrate the capabilities of our approach. One notable strength of our method is its ability to creatively combine basic shapes, such as circles, rectangles, and rounded rectangles, to produce sophisticated layouts. 
These compositions often exhibit visual qualities and structural coherence that cannot be achieved by merely placing contents (text and images). 
By training the model to generate simple elements, our approach generates visually appealing and professionally cohesive designs, demonstrating an understanding of complex design principles such as alignment, symmetry, and spatial balance. 
This capability significantly surpasses simpler layout generation methods, underscoring the practical utility of our iterative, refinement-based framework.

\end{document}